\documentclass[2]{sig-alternate}
\usepackage{graphicx}
\usepackage{amssymb}
\usepackage{times}
\usepackage{verbatim}
\usepackage{amsmath}
\usepackage{color}
\usepackage[vlined,english,ruled]{algorithm2e}
\usepackage{import}

\newcommand{\set}[1]{\left\{{#1}\right\}}
\newcommand{\range}[2]{\set{{#1},\dots,{#2}}}

\graphicspath{{figures/}}

\renewcommand{\S}[0]{{\mathcal S}}

\begin{document}
\conferenceinfo{GECCO'11,} {July 12--16, 2011, Dublin, Ireland.}
\CopyrightYear{2011}
\crdata{978-1-4503-0557-0/11/07}
\clubpenalty=10000
\widowpenalty = 10000

\title{DAMS: Distributed Adaptive Metaheuristic Selection}
\date{}

\numberofauthors{2}

\author{
\alignauthor
Bilel Derbel\\
       \affaddr{Universit\'e Lille 1}\\
       \affaddr{LIFL -- CNRS -- INRIA Lille}\\
       \email{bilel.derbel@lifl.fr}
\alignauthor
S\'ebastien Verel\\
       \affaddr{Univ. Nice Sophia-Antipolis}\\
       \affaddr{INRIA Lille}\\
       \email{verel@i3s.unice.fr}
}

\maketitle
\begin{abstract}
We present a distributed algorithm, Select Best and Mutate (SBM), in the Distributed Adaptive Metaheuristic Selection (DAMS) framework. DAMS is dedicated to adaptive optimization in distributed environments. Given a set of metaheuristics, the goal of DAMS is to coordinate their local execution on distributed nodes in order to optimize the global performance of the distributed system. DAMS is based on three-layer architecture allowing nodes to decide distributively what local information to communicate, and what metaheuristic to apply while the optimization process is in progress. SBM is a simple, yet efficient, adaptive distributed algorithm using an exploitation component allowing nodes to select the metaheuristic with the best locally observed performance, and an exploration component allowing nodes to detect the metaheuristic with the actual best performance. SBM features are analyzed from both a parallel and an adaptive point of view, and its efficiency is demonstrated through experimentations and comparisons with other adaptive strategies (sequential and distributed).
\end{abstract}





\section{Introduction}
\label{sec:intro}

\textbf{Motivation:} Evolutionary algorithms or metaheuristics are efficient stochastic methods
for solving a wide range of optimization problems.
Their performances are often subject to a correct setting of their parameters
including the representation of solution,
the stochastic operators such as mutation, or crossover, 
the selection operators,
the rate of application of those operators,
the stopping criterium, or
the population size, etc.
From the metaheuristics-user point of view,
it could be a real challenge to choose the parameters, and understand this choice.
Moreover, when the algorithms evolve in a distributed environment,
comes additional possible designs, or parameters such as
the communication between distributed nodes,
the migration policy, 
the distribution of the metaheuristics execution, 
or when different metaheuristics are used together, their distribution over the environement, etc.
The aim of this paper is the parameter setting in a fully distributed environment.

\textbf{Background and related work:} Following the taxonomy of Eiben et al. \cite{EibenMSS07},
we distinguish two types of parameter setting:
the first one is \textit{off-line}, before the actual run, often called \textit{parameter tuning},
and the second one is \textit{on-line}, during the run, called \textit{parameter control}.
Usually, parameter tuning is done by testing all or a subset of possible parameters, 
and select the combination of parameters which gives the best performances.
Obviously, this method is time consuming. 
Besides, a parameters setting at the beginning of the optimization process could be inappropriate at the end.
Several approaches are used for the parameters control.
Determinist ones choose an \textit{a-priori} policy of modification.
But, the choice of this policy gets more complicated compared to the static setting as the complexity of design increases.
Self-adaptive techniques encode the parameters into the solution and evolve them together.
This approach is successfully applied in continuous optimization \cite{Hansen01}.
Nevertheless, in combinatorial optimization it often creates a larger search space leading to efficiency loss.
Finally, adaptive methods use the information from the history of the search to modify the parameters.
The hyperheuristics are one example \cite{Burke03}. They are heuristics that adaptively control (select, combine, generate, or adapt) other heuristics. Another example is the Adaptive Operator Selection (AOS) which controls the application of the variation operator, using probability matching, adaptive pursuit \cite{Thierens:2005}, or multi-armed bandit \cite{Fialho08} techniques. In this work, we focus on parameters control methods in a distributed setting.

In fact, the increasing number of CPU cores, parallel machines like GPGPU, grids, etc.
makes more and more the distributed environments the natural framework to design effective optimization algorithms.
Parallel Evolutionary Algorithms (EA) are well suited to distributed environments, 
and have a long history with many success \cite{Tomassini:2005}.
In parallel EA, the population is structured: 
the individuals interact only with their neighbor individuals.
Two main parallel EA models can be identified.
In the island (or multi-population) one, the whole population is divided into several ones.
In the cellular model, the population is embedded on a regular toroidal grid.
Of course, between this coarse grained to fined grained, many variant exist.
This field of research received more and more attention.
Recent theoretical and experimental works study the influence of parallelism \cite{Lassig:2010:BMP:1830483.1830687,Lassig10},
or study the fined grained population structure on performances \cite{Laredo2011}.

Previous works control the specific parameters of parallel EA, or others parameters.
In \cite{Bonnaire05}, the architecture of workers with a global controller is used to self-adapt the population size in each island. Population size, and number of crossover points is controlled using the average value in the whole population in \cite{Srinivasa:2007}. 
Those two examples of works use a parallel environment where a global information can be shared.
In the work of Tongchim et al. \cite{Tongchim:2002}, a distributed environment is considered for the parameters control.
Each island embeds two parameters settings. Each parameters is evaluated on half the population, 
and the best parameter setting is communicated to the other islands which is used to produce new parameters.
So, parameters are controlled in a self-adaptive way using local comparison between two settings only.
Laredo et al. \cite{Laredo:2010} propose the gossiping protocol newcast for P2P distributed population EA.
The communication between individuals evolves towards small-world networks. Thus, the selection pressure behaves like in panmictic population, but population diversity and system scalability outperform the fully connected population structure.

In summary, four main issues has been studied in parameter control in parallel EA:
static strategies where the parameters can be different for each island but does not change during the search;
strategies that use a global controller;
self-adaptive strategies where each node compares possible parameter settings;
and distributed strategies that evolve the communication between nodes.
In this work, we propose a new and intuitive way to control the parameters in distributed environment:
during the search, 
from a set of metaheuristics which correspond to possible parameter settings,
each node select one metaheuristic to execute according to local information (\textit{e.g.} a performance measure) given by the other nodes.

\textbf{Contributions and outline:} In this paper, we define a new adaptive parameters control method called distributed adaptive metaheuristic selection (DAMS) dedicated to distributed environments. DAMS is designed in the manner of heterogeneous island model EAs where different metaheuristics live together. 
In this context,
the distributed strategy (Select Best and Mutate strategy) which selects either the best metaheuristic identified locally, or one random metaheuristic is defined and studied.
For more simplicity, but also to enlighten the main features of this strategy, the SBM is given in the framework of DAMS.
Generally speaking, from an external observer, a distributed environment can be viewed as a unique global system which can be optimized in order to perform efficiently. However, the existence of such a global observer is not possible in practice nor mandatory. From DAMS point of view, the optimization process in such a distributed environment is though in a very local manner using only local information to coordinate the global distributed search and guide the optimization process. 
The issue of metaheuristic selection in distributed environments is introduced in Section~\ref{sec:DAMS}. 
From that issue, the DAMS framework is given in Section~\ref{sec:dams} bringing out three levels of design.
The select Best and Mutate strategy (SBM) is then addressed in Section~\ref{sec:simpleDAMS}. Different SBM properties are studied in Section~\ref{sec:expDAMS} through extensive experimentations using the oneMax problem. Finally, in Section~\ref{sec:conc} we conclude with some open issues raised by DAMS.

\section{Selection of metaheuristic in distributed environments}
\label{sec:DAMS}

\subsection{Position}
Let us assume that to solve a given optimization problem, we can use a set $\mathcal{M}$ of \emph{atomic function} that we can apply in an \emph{iterative} way on a solution or a population of solutions. 
By atomic function we mean a black-box having well defined input and output specifications corresponding to the problem being solved. In other words, we do not have any control on an atomic function once its is executed on some input population of solutions. To give a concrete example, an atomic function could be a simple (deterministic or stochastic) move according to a well defined neighborhood. It could also consist in applying 
the execution of a metaheuristic for fixed number of iterations. Hence, an atomic function can be viewed from a very general perspective as a undividable, fixed and fully defined metaheuristic.
So, in the following we call an atomic metaheuristic, or shorter a metaheuristic, the atomic functions.

In this context, 
where the atomic metaheuristics can be used in a sequential way, 
designing a strategy to solve an optimization problem turns out to design an iterative algorithm that carefully combines the atomic functions (metaheuristics) in a specific order. In this paper, we additionally assume that we are given a set of computational resources that can be used to run the atomic metaheuristics. These computational resources could be for instance a set of physical machines distributed over a network and exchanging messages, or some parallel processors having some shared memory to communicate.

Having such a distributed environment and the set of atomic metaheuristic at hand, the question we are trying to study in this paper is as following: \emph{how can we design an efficient distributed strategy to solve the optimization problem?} Obviously, the performance of the designed strategy depends on how the \emph{distributed} combination of the atomic operations is done. In particular, one have to \emph{adapt} the search at runtime and decide \emph{distributively} which atomic function should be applied at which time by which computational entity. The concept of DAMS introduced in this paper aims at defining in a simple way a general algorithmic framework allowing us to tackle the latter question. 

\subsection{Example of an optimal strategy}

\begin{figure}[ht!]
\begin{center}
\includegraphics[width=0.45\textwidth]{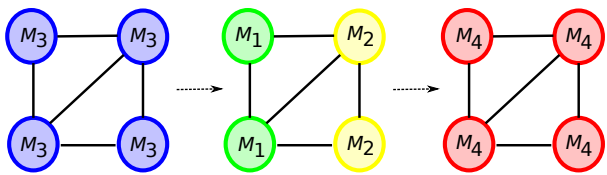}  
\caption{Illustration of the distributed oracle strategy of algorithm~\ref{algo:example}. 
At the beginning all nodes $\range{n_1}{n_4}$ run $M_3$, then $M_1$ and $M_2$ are used, and finally $M_4$ concludes the distributed search.
\label{fig:example}}
\end{center}
\end{figure}

To give an algorithm following the DAMS framework, let us consider the following simple example. Assume we have a network of $4$ nodes denoted by: $\range{n_1}{n_4}$. Any two nodes, but nodes $n_1$ and $n_4$ (see Fig~\ref{fig:example}), can communicate together by sending and receiving messages throughout the network. Assume that the atomic operations we are given are actually composed of four population-solution based metaheuristics denoted by: $\range{n_1}{n_4}$, i.e., each metaheuristic accepts as input a population of solutions and outputs a new one. Then, starting with a randomly generated population, our goal is to design a distributed strategy that decides how to distribute the execution of our metaheuristics on the network and what kind of information should be exchanged by computational nodes to obtain the best possible solution. To make it simple, assume that the best strategy, which can be considered as a distributed oracle strategy, is actually given by distributed algorithm~\ref{algo:example} (See also Fig.~\ref{fig:example} for an illustration). In other words, any other distributed strategy cannot outperforms Algorithm~$1$. 

\begin{algorithm}
Generate a random solution on each node $n_i$ with $i\in \range{1}{4}$\;
Run $M_3$ in parallel on each $n_i$ with $i\in \range{1}{4}$\;
Exchange best found solution\;
Run $M_1$ in parallel on each $n_i$ with $i\in \set{1,2}$\;
Run $M_2$ in parallel on each $n_i$ with $i\in \set{3,4}$\;
Exchange best found solution\;
Run $M_4$ in parallel on each $n_i$ with $i\in \range{1}{4}$\;
Return best found solution\;
\caption{Oracle strategy of distributed adaptive metaheuristic selection}\label{algo:example}
\end{algorithm}

Having this example in mind, the question we are trying to answer is then: How can we design a distributed strategy which is competitive compared to the oracle of algorithm~\ref{algo:example}? 
Finding such a strategy is obviously a difficult task. In fact, finding the best mapping of our metaheuristics into the distributed environment is in itself an optimization problem which could be even harder than the initial optimization problem we are trying to solve. In this context, one possible solution is to map the metaheuristics into the distributed environment in an adaptive way in order to guide the distributed search and operate as close as possible of the oracle strategy (Algorithm~\ref{algo:example} in our example). This is exactly the ultimate goal of the adaptive metaheuristic selection in distributed environments. In the following section, we introduce a general algorithm, DAMS, and discuss its different components.

\section{DAMS: a generic framework}
\label{sec:dams}

Generally speaking, a Distributed Adaptive Metaheuristic Selection (DAMS) is an adaptive strategy that allows computational nodes to coordinate their actions distributively by exchanging local information. Based only on a local information, computational nodes should be able to make efficient local decisions that allows them to guide the global search efficiently (i.e., to be as efficient as possible compared to a distributed oracle strategy).

Before going into further details and for the sake of simplicity, we shall abstract away the nature of the distributed environment under which a DAMS will be effectively implemented. For that purpose, we use a simple unweighted graph $G=(V,E)$ to model the distributed environment. A node $v\in V$ models a computational node (e.g., a machine, a processor, a process, etc). An edge $(u,v) \in E$ models a bidirectional direct link allowing neighboring nodes $u$ and $v$ to communicate together (e.g., by sending a message through a physical network, by writing in a shared local/distant memory, etc).

A high level overview of DAMS is given in distributed algorithm~\ref{algo:DAMS}. Notice that the high level code given in algorithm~\ref{algo:DAMS} is executed (in parallel) by each node $v\in V$. The input of each node $v$ is a set $\mathcal{M}$ of atomic metaheuristics.

\begin{algorithm}
\KwIn{A set of metaheuristics $\mathcal{M}$}
$M \leftarrow \textsc{Init\_Meta()}$\;
$P\leftarrow \textsc{Init\_Pop()}$\; 
$\S\leftarrow \textsc{Init\_State()}$\; 
$I \leftarrow \{ M, P, \S \}$; \CommentSty{/$^{*}$$v$'s initial local information$^{*}$/}\\ 
\Repeat{\textsc{stopping\_condition}$(I)$}{
	\CommentSty{/$^{**}$ Distributed Level $^{**}$/}\\
	$c \leftarrow$ \textsc{Local\_Communication}$(I)$\;
       $P \leftarrow$ \textsc{Update\_Population}$(I, c)$\;
	$\S \leftarrow$ \textsc{Update\_Local\_State}$(I, c)$\;
	\CommentSty{/$^{**}$ Metaheuristic Selection Level $^{**}$/}\\
	$M \leftarrow \textsc{Select\_Meta}(I)$\;
	\CommentSty{/$^{**}$ Atomic Low Level $^{**}$/}\\
       $(P,\S) \leftarrow$ \textsc{Apply\_Meta}$(M, P)$\;
}
\caption{{\em DAMS}: high level code for node $v\in V$} \label{algo:DAMS}
\end{algorithm}

Each node $v$ participating into the computations has three \emph{local} variables: A population $P$ which encodes a set of individuals. A set $\S$ which encodes the local state of node $v$. The local state $\S$ of node $v$ is updated at each computation step and allows node $v$ to make local decisions. Finally, each node has a current metaheuristic denoted by $M$. 

As depicted in algorithm~\ref{algo:DAMS}, a DAMS operates in many rounds until some stopping condition is satisfied. Each round is organized in three levels.

\textbf{The Distributed Level}: This level allows nodes to communicate together according to some distributed scheme and to update their local information. This step is guided by the current local information $I=\set{M, P, \S}$ which encodes the experience of node $v$ during the distributed search. Typically, a node can sends and/or receives a message to one or many of its neighbors to share some information about the ongoing optimization process. According to its own local information (variable $I$) and the local information exchanged with neighbors (variable $c$), node $v$ can both update its local state $\S$ and its local population $P$. Updating local state $\S$ is intended to allow node $v$ to record the experience of its neighbors for future rounds and make local decisions accordingly. Updating local population $P$ aims typically at allowing neighboring nodes to share some representative solutions found so far during the search, e.g. best individuals. This is the standard migration stage of parallel EA. Finally, we remark that since at each round local information $I$ of node $v$ is updated with the experience of its neighbors, then after some rounds of execution, we may end with a local information $I$ that reflects the search experience of \emph{not} only $v$'s neighbors but of other nodes being farther away in the network graph $G$, e.g., the best found solution or the best metaheuristic could be spread through the network with a broadcast communication process.

\textbf{The Metaheuristic Selection Level}: This level allows a node to locally select a specific metaheuristic $M$ to apply from set $\mathcal{M}$ of available metaheuristics. First, we remark that this level may \emph{not} be independent of the distributed level. In fact, the decision of selecting a metaheuristic is guided by local information $I$ which is updated at the distributed level. Although the metaheuristic choice is executed locally and independently by each node $v$, it may \emph{not} be independent of other nodes choices. In fact, the local communication step shall allow nodes to coordinate their decisions and define distributively a cooperative strategy. Second, since at the distributed level, local information $I$ of a a given node $v$ could store information about the experience of other nodes in the network, the metaheuristic choice can clearly be adapted accordingly. Therefore, the main challenge when designing an efficient DAMS is to coordinate the Metaheuristic Selection level and the Distributed Level in order to define the best distributed optimization strategy.

\textbf{The Atomic Low Level}:  At this level a node simply executes the selected metaheuristic $M$ using its population $P$ as input. Notice that 'atomic' refers to the fact that a node cannot control the sequential metaheuristic execution in any way. However, after applying a given metaheuristic, the population and the local state of each node may be updated. For instance, one can decide to move to the new population following a given criterion or to simply record some information about the quality of the outputted population using local state $\S$.

\section{Select Best and Mutate strategy}
\label{sec:simpleDAMS}

The originality of DAMS framework is to introduce a metaheuristic selection level which adaptively select a metaheuristic according to local information shared by the other neighboring nodes.
In this section, we give a metaheuristic selection strategy, the Select Best and Mutate strategy (SBM), which specify the DAMS.

For the sake of clarity, we consider the classical message passing model: We consider an $n$-node simple graph $G=(V,E)$ to model a distributed network. Two nodes $u$ and $v$, such that $(u,v)\in E$, can communicate together by sending and receiving messages toward edge $(u,v)$. A high level description of SBM is given in algorithm~\ref{algo:SBM} and is discussed in next paragraphs. Note that the code of algorihtm~\ref{algo:SBM} is to be executed (in parallel) on every node $v\in V$. The algorithm accepts as input a finite set of  metaheuristics $\mathcal{M}=\cup_{k=1}^m \{ M_k \}$ and a probability parameter $p_{mut}$.

\begin{algorithm}[ht!]
\SetKwInput{KwIn}{\textbf{Inputs}}
\KwIn{A set of metaheuristics $\mathcal{M}=\cup_{k=1}^m \{ M_k \}$;\\ ~~~~~~~~~~~~~~A metaheuristic mutation rate $p_{mut} \in [0,1] $ \;}
$r \leftarrow 0$\;
$k \leftarrow \textsc{Init\_Meta}(\mathcal{M})$\;
$P \leftarrow \textsc{Init\_Pop()}$\;
\Repeat{Stopping condition is satisfied}{
	\CommentSty{/$^*$ Distributed Level $^*$/}\\
	Send \textsf{Msg($r,k,P$)} to each neighbor\;
	$\mathcal{P} \leftarrow \set{P}$; $\mathcal{S} \leftarrow \set{(r,k)}$\; 
	\For{each neighbor $w$}{
		Receive \textsf{Msg($r',k',P'$)} from $w$\;
		$\mathcal{P} \leftarrow \mathcal{P} \cup P'$\;
		$\mathcal{S} \leftarrow \mathcal{S} \cup \set{(r',k')}$\;		
	}
	$P \leftarrow \textsc{Update\_Population}(\mathcal{P})$\;
	\CommentSty{/$^*$ Metaheuristic Selection Level $^*$/}\\
	$k_{best} \leftarrow \textsc{Select\_Best\_Meta}(\mathcal{S})$\; 
	\If{$\textsc{Rnd}(0,1) < p_{mut}$}{
		$k \leftarrow \textsc{Rnd}(\mathcal{M} \setminus M_{k_{best}})$
	}\lElse{
	$k \leftarrow k_{best}$\;
	}
	\CommentSty{/$^*$ Atomic Low Level $^*$/}\\
       	Find a new population $P_{new}$ by applying metaheuristic $M_{k}$ with $P$ as an initial population\;
	$r\leftarrow \textsc{Reward}(P,P_{new})$\;
	$P\leftarrow P_{new}$\;
}

\caption{SBM code for every node $v\in V$}
\label{algo:SBM}
\end{algorithm}

Clearly, SBM follows the general scheme of a DAMS defined previously in Algorithm~\ref{algo:DAMS}. One can distinguish the three basic levels of a DAMS and remark their inter-dependency.\\

\textbf{SBM Distributed level}: The local information defined by a DAMS is encoded implicitly in SBM using variables $r$, $k$ and $P$. For every node $v$, variable $k$ refers to metaheuristic $M_k$ considered currently by $v$. Variable $r$ (reward) refers to the quality of metaheuristic $M_k$. Variable $P$ denotes the current population of node $v$. SBM operates in many rounds until a stopping condition is satisfied. At each round, the distributed level consist in sending a message containing triple $(r,k,P)$ to neighbors and symmetrically receiving the respective triples $(r',k',P')$ sent by neighbors. Then after, node $v$ constructs two sets $\mathcal{P}$ and $\S$. Set $\mathcal{P}$ contains information about neighbors populations. Node $v$ can then update its current population $P$ according to set $\mathcal{P}$. For instance, one may selects the best received individuals, or even apply some adaptive strategy taking into account the search history. As for set  $\S$, it gathers information about the quality of other metaheuristics considered by neighbors. This set is used at the next DAMS level to decide on the new current metaheuristic to be chosen by node $v$.\\

\textbf{SBM Metaheuristic Selection Level}: choosing a new metaheuristic is guided by two ingredients. Firstly, using set $\S$, node $v$ selects a metaheuristic according to neighbors information. One can imagine many strategies for selecting a metaheuristic possibly depending not only on $\S$ but also on received populations $\mathcal{P}$. In SBM, we simply select metaheuristic $k_{best}$ corresponding to the best received reward $r$. Notice that defining what is the best observed metaheuristic could be guided by different policies. From an exploitation point of view, function $\textsc{Select\_Best\_Meta}$ has the effect of pushing nodes to execute the metaheuristic with the best observed performance during the search process. Obviously, a metaheuristic with good performance at some round could quickly becomes inefficient as the search progresses. To control this issue, we introduce a \emph{exploration} component in the selection level. In SBM, this component is simply guided by a metaheuristic mutation operator. In fact, every node decides to select at random another metaheuristic different from $M_{k_{best}}$ with rate $p_{mut}$. Intuitively, these two ingredients in SBM selection level shall allow distributed nodes to obtain a good tradeoff between exploitation and exploration of metaheuristics, and adapt their decisions according to the search.\\

\textbf{SBM Atomic Low Level}: Once a new metaheuristic $M_{k}$ is selected, it is used to compute a new population. To evaluate the performance of metaheuristic $M_k$, SBM simply compares the previous population $P$ with the new population $P_{new}$ using a generic $\textsc{Reward}$ function. For instance, a simple evaluation strategy could be by comparing the best individual fitness, the average population fitness, or even more sophisticated adaptive strategies taking into account the performances observed in previous rounds.

\section{Experimental study of SBM}
\label{sec:expDAMS}
In this section, we study a specific SBM-DAMS by fully specifying its different levels. We report the results we have obtained by conducting experimental campaigns using the oneMax problem.
\subsection{SBM setting}

To conduct experimental study of the SBM-DAMS, we test the algorithm on a classical problem in EA, the \textit{OneMax} problem. This problem was used in recent works of Fialho et al. \cite{Fialho10} which propose a (sequential) Adaptive Operator Selection (AOS) method based on the Multi-Armed Bandits, and a credit assignment method which uses the fitness comparison. The oneMax problem, the "drosophila" of evolutionary computation, is a unimodal problem defined on binary strings of size $l$. The fitness is the number of "$1$" in the bit-string. Within the AOS framework, the authors in~\cite{Fialho10} considered $(1+\lambda)-$EA and four mutation operators to validate their approach: The standard $1/l$ bit-flip operator (every bit is flipped with the binomial distribution of parameter $1/l$ where $l$ is the length of the bit-strings), and the $1-$bit, $3-$bit, and $5-$bit mutation operators (the $b-$bit mutation flips exactly $b$ bits, uniformly selected in the parents). In the rest of the paper, we shall also consider the same four atomic metaheuristics to study DAMS. Each atomic function is one iteration of $(1+\lambda)$-EA using one of the four mutation operators. Unless stated explicitly, parameter $\lambda$ is set to $50$ as in \cite{Fialho10}. Notice also by studying SBM-DAMS with these well-understood operators does not undergo any major weakness of our approach since these operators mainly exhibits different exploration degrees that one can encounter in other settings when using other operators.

In all reported experiments, the length of the bit strings is set to $l=10^4$ as in \cite{Fialho10}. Population $P$ of SBM-DAMS is reduced to a single solution $x$. The initial solution is set to $(0, \ldots, 0)$. The rewards $r$ is the fitness gain between parent and offspring solutions: $f(x_{new}) - f(x)$. The migration policy (\textsc{Update\_Population}) is elitist, it replaces the current solution by one of the best received solutions if its fitness is strictly higher. The algorithm stops when the maximal fitness $l$ is reached. This allows us to compare the performances of SBM-DAMS according to the number of evaluations as used in sequential algorithms; but also in terms of rounds (total number of migration exchange) as used in parallel frameworks, and in terms of messages cost that is the total number of local communications made through the distributed environment.

In the remainder, the \textit{oneMax} experimental protocol is used to first compare the efficiency of SBM-DAMS to some oracle and naive strategies, then the parallel properties are analyzed with a comparison to the sequential AOS method.



\subsection{Experimental setup}
\label{sec:setup}

Three network topologies are studied: complete, grid, and cycle. In the complete one, the graph is a clique, i.e., every node is linked to all other nodes. In the grid topology, every node can communicate with four other neighbors except at the edge of the grid (non-toroidal grid). In the circle topology, nodes are linked to two others nodes to form a circle.

Since there exist no previous \emph{distributed} approaches addressing the same issues than DAMS, we shall compare SBM-DAMS to the state-of-the-art sequential adaptive approaches (Section~\ref{sec:parallelism}) and also to two simple distributed strategies (Section~\ref{sec:adaptation}) called rnd-DAMS and seqOracle-DAMS. (i)~In rnd-DAMS, each node selects at random (independently of node states) at each round a metaheuristic to be executed. It allows us to evaluate the efficiency of SBM selection method based on the best instant rewards. (ii)~In seqOracle-DAMS, each node executes the sequential oracle which gives the operator with the maximum fitness gain according to the fitness of node current solution. The sequential oracle is taken from the paper \cite{Fialho08}. It is used independently by each node without taking into account rewards, or observed performances of others nodes. Notice that that seqOracle-DAMS outperforms a static strategy which use only one metaheuristic, but it is different from a pure distributed oracle which could lead to better performances for the whole distributed system, by taking take into account all nodes to select the metaheuristic of each node.

For the latter two based-line strategies, we use the same elitism migration policy as SBM-DAMS and evaluate their performances using the same three topologies. For each topology, network size is $n \in \{4, 8, 16, 36, 64\}$, and the performance measures are reported over $20$ independent runs. Notice that for this particular study, the number of evaluations is $\lambda . n$ times the number of rounds. The number of exchanged messages is $|E|$ times the number of rounds where $|E|$ is the number of communication links in the considered topology.

\subsection{Adaptation properties}
\label{sec:adaptation}

\begin{figure}[ht!]
\begin{center}
\begin{tabular}{c}
\includegraphics[width=0.3\textwidth]{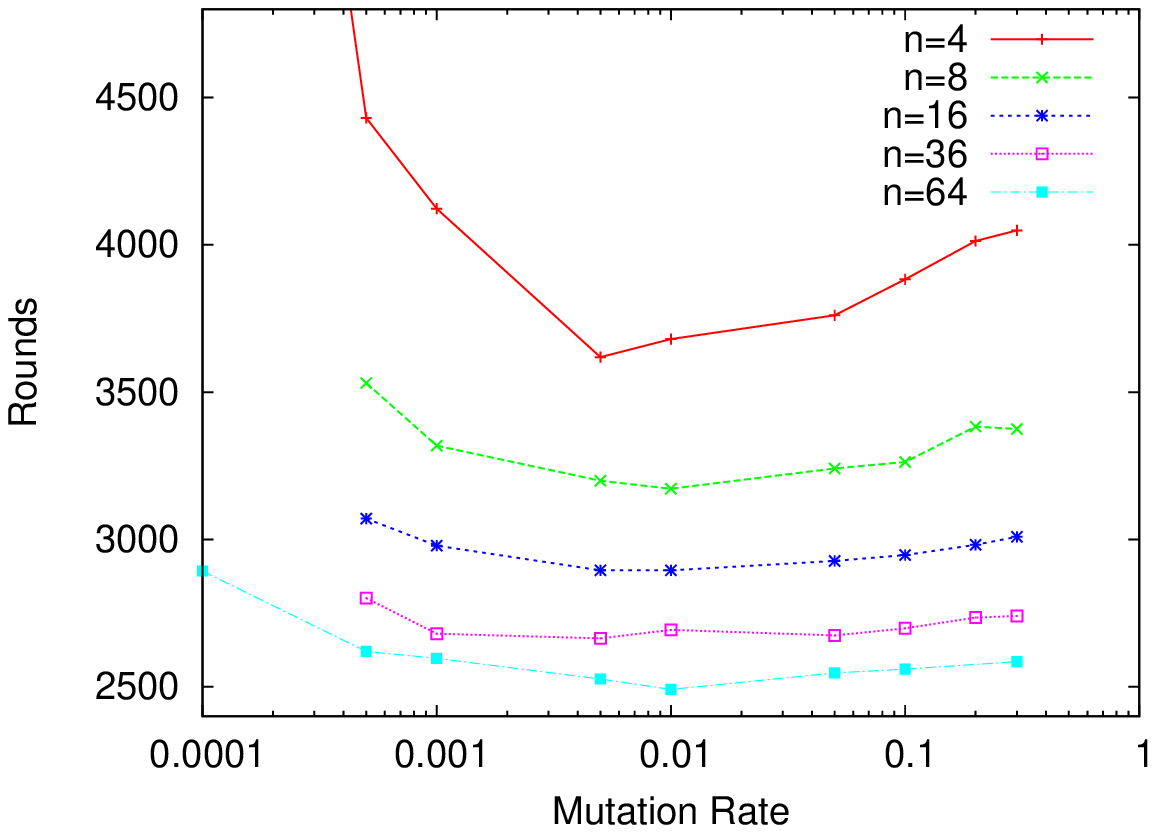}\\
\includegraphics[width=0.3\textwidth]{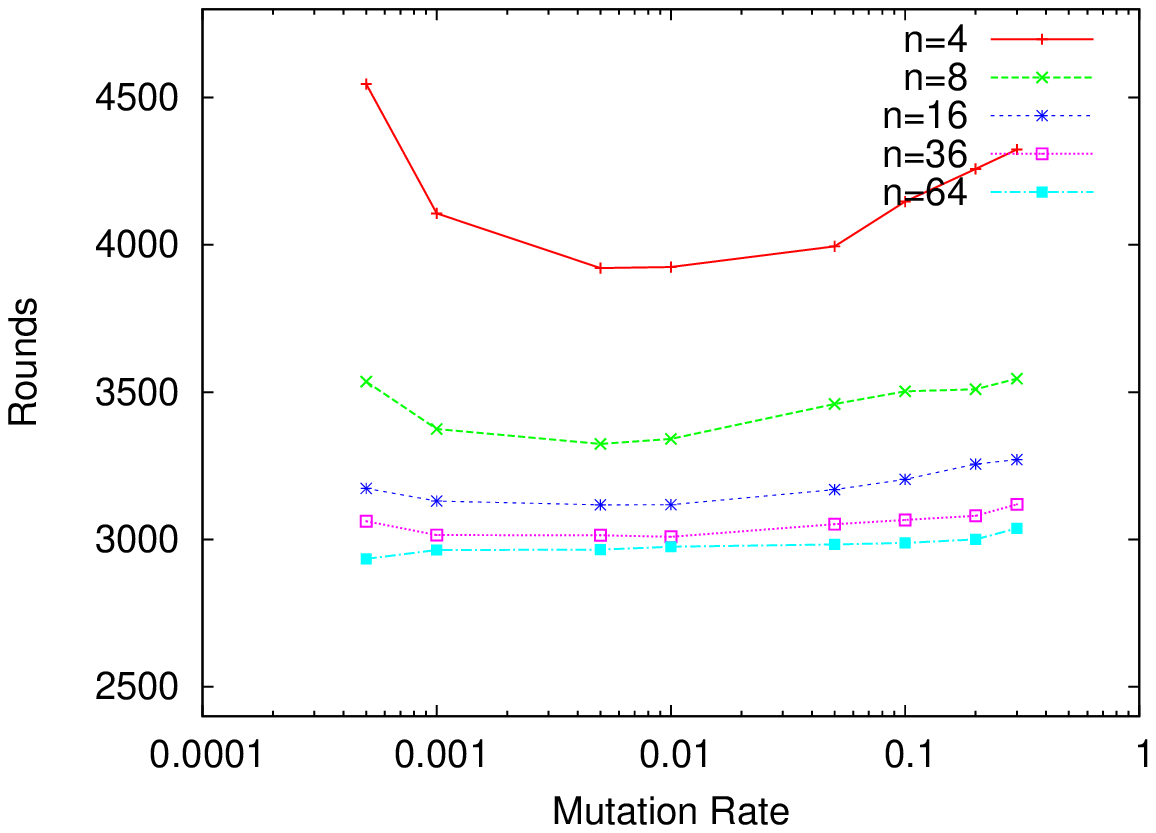} \\
\includegraphics[width=0.3\textwidth]{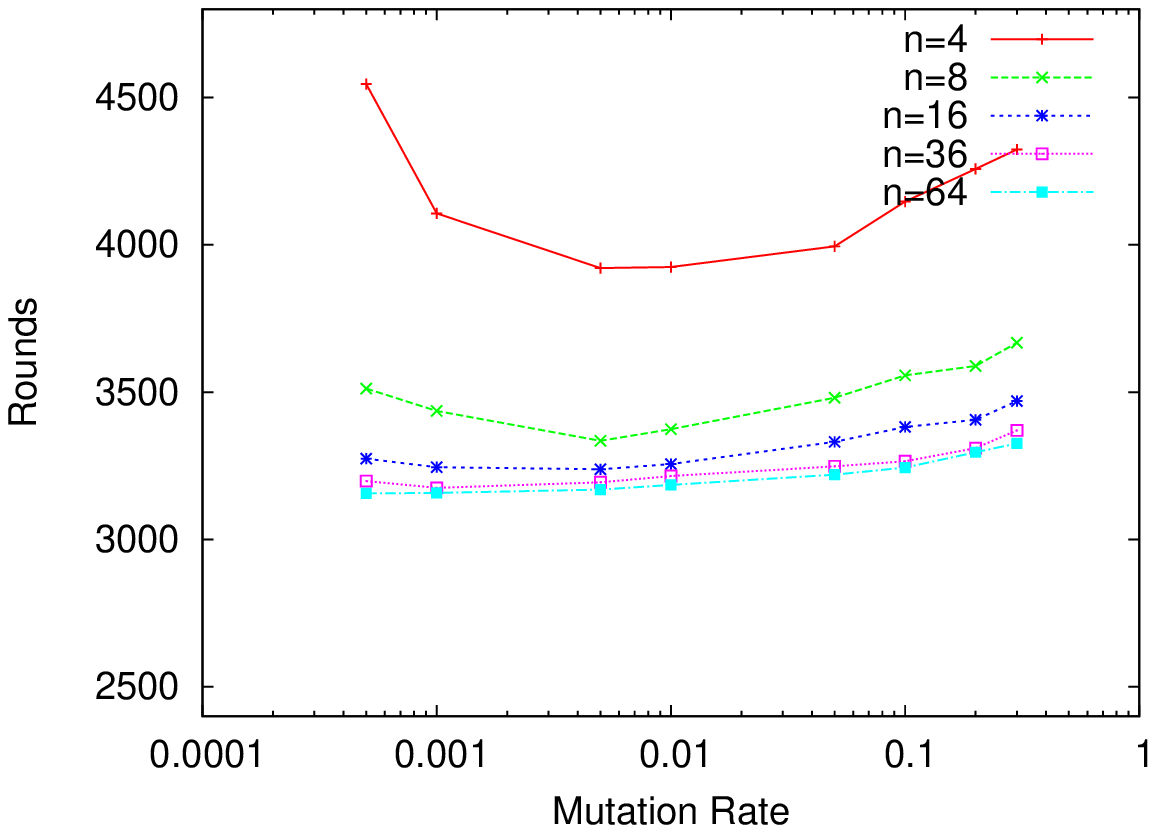} \\
\end{tabular}
\caption{Average number of rounds
according to metaheuristic mutation rate
for different sizes $n$ and different topologies (from top to bottom: complete, grid, and circle).}
\label{fig:rnd-mutation}
\end{center}
\end{figure}

To tune off-line, the metaheuristics mutation rate of SBM-DAMS,
we perform a design of experiments campain with $p_{mut} \in \{ 0.0005,$ $0.005, 0.001, 0.01, 0.05, 0.1, 0.2, 0.3 \}$.
Fig.~\ref{fig:rnd-mutation} shows the average number of rounds 
to reach the optimum according to mutation rate $p_{mut}$ for different topologies and network sizes.
Except for small size $n=4$, the range of performances is small according to the mutation rate.

\begin{figure}[ht!]
\begin{center}
\begin{tabular}{c}
\includegraphics[width=0.3\textwidth]{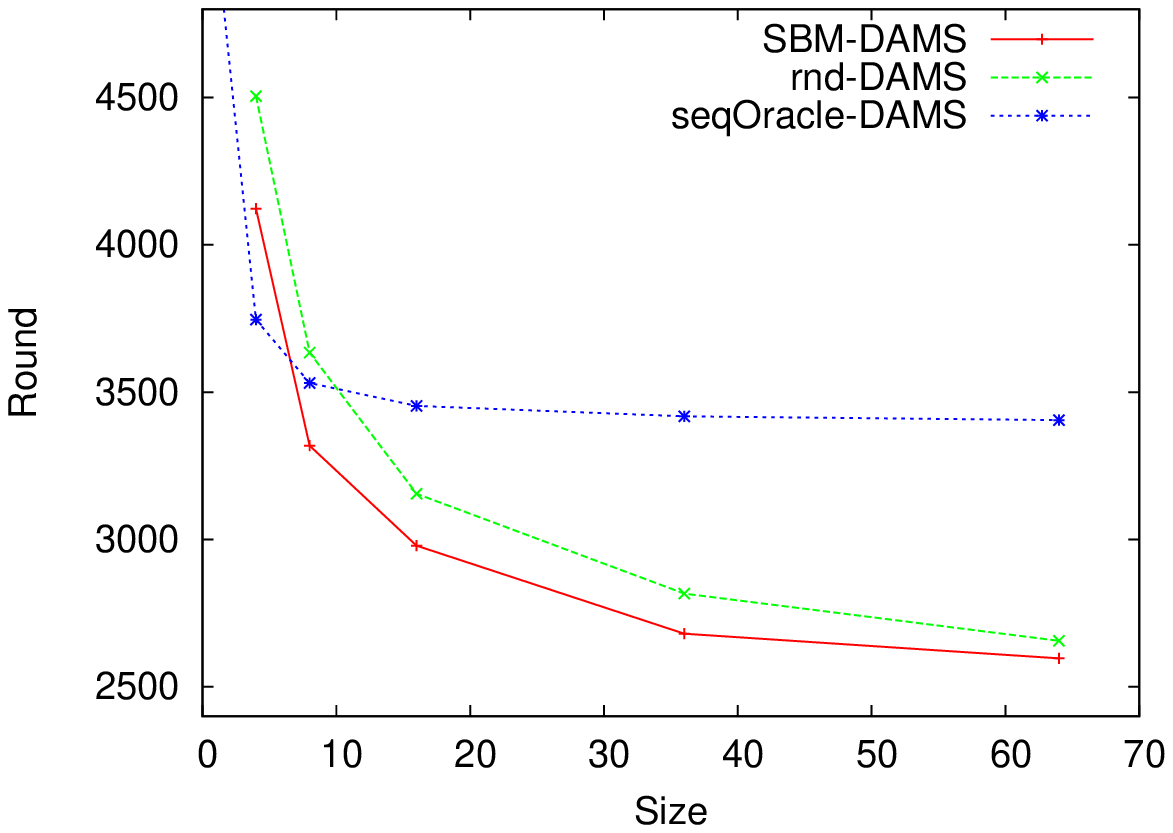} \\ 
\includegraphics[width=0.3\textwidth]{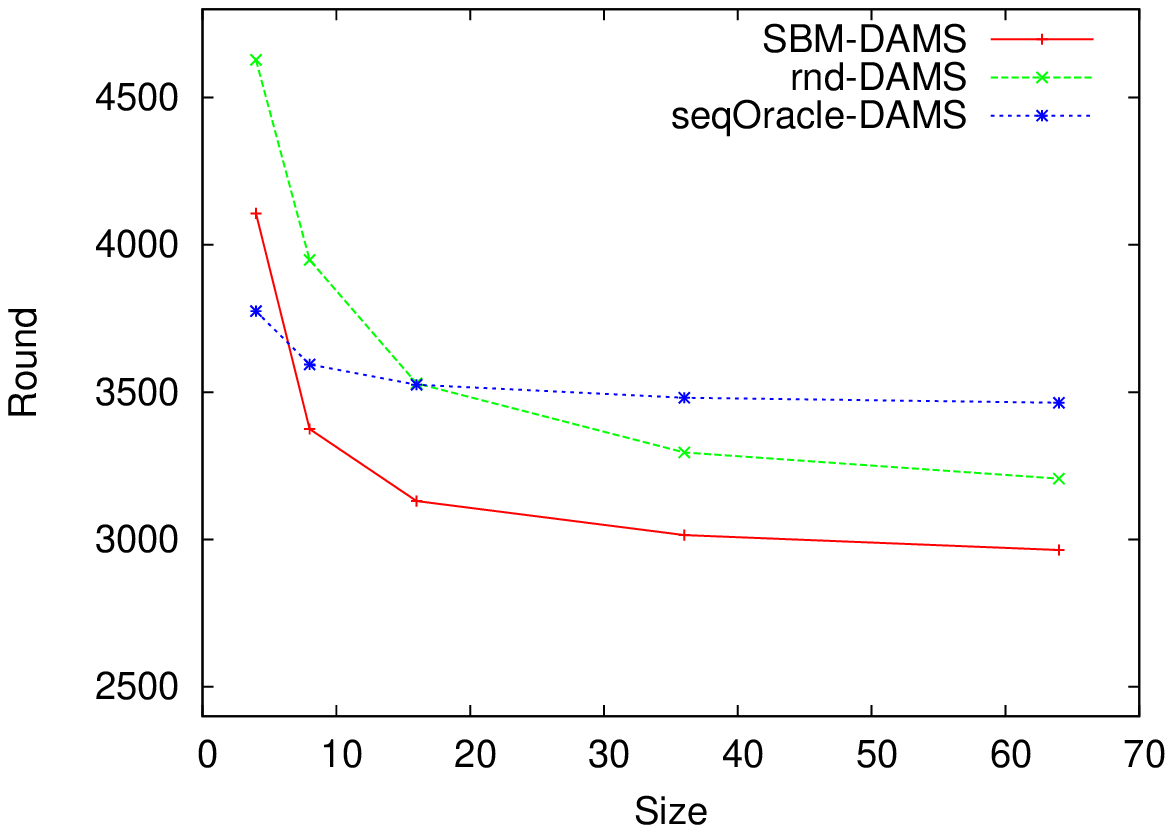} \\
\includegraphics[width=0.3\textwidth]{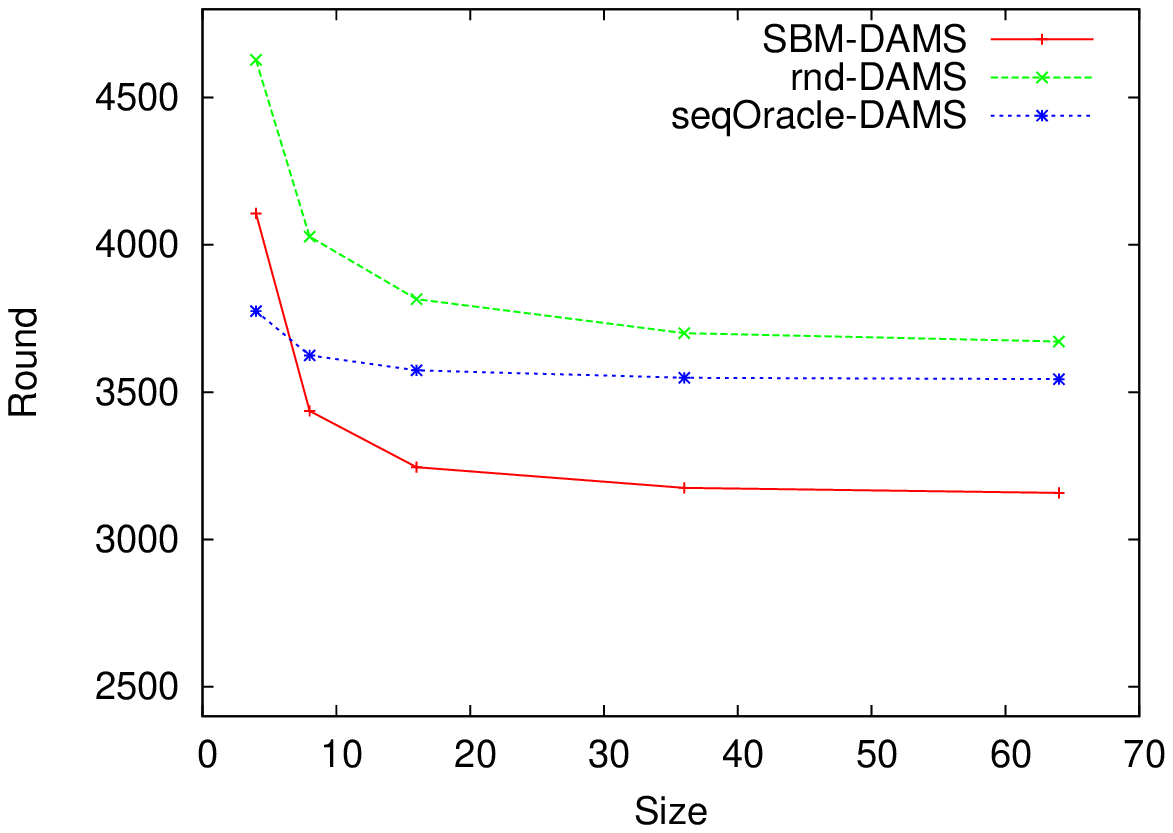} \\
\end{tabular}
\caption{Average number of rounds
according to size of the network
for different topologies (from top to bottom: complete, grid, and circle)}
\label{fig:round-Size}
\end{center}
\end{figure}

Fig.~\ref{fig:round-Size} shows the average number of rounds to reach the optimum 
for different DAMS as a function of network size and according to the three different topologies.
The metaheuristic mutation rate is set to $p_{mut} = 10^{-3}$ for SBM-DAMS. 
SBM-DAMS outperforms the based-line strategies regardless to topology, except for small networks with only $4$ nodes. 
The difference is statistically significant according to the non-parametric test of Mann-Whitney at $1 \%$ level of confidence.
This shows that the selection best-mutation scheme of SBM-DAMS is efficient against other naive strategies.

\begin{figure}[ht!]
\begin{center}
\begin{tabular}{ccc}
\hspace{-9ex}\includegraphics[width=0.3\textwidth]{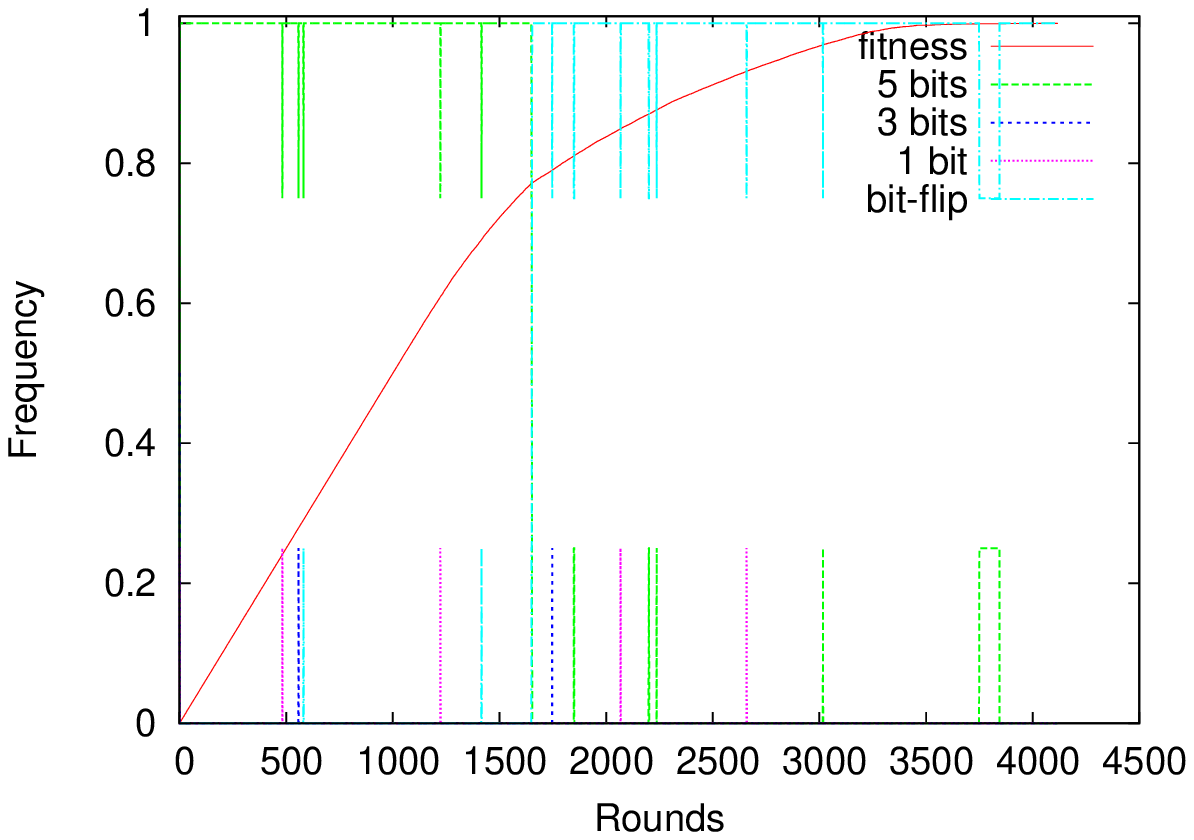} & \hspace{-5ex}\includegraphics[width=0.3\textwidth]{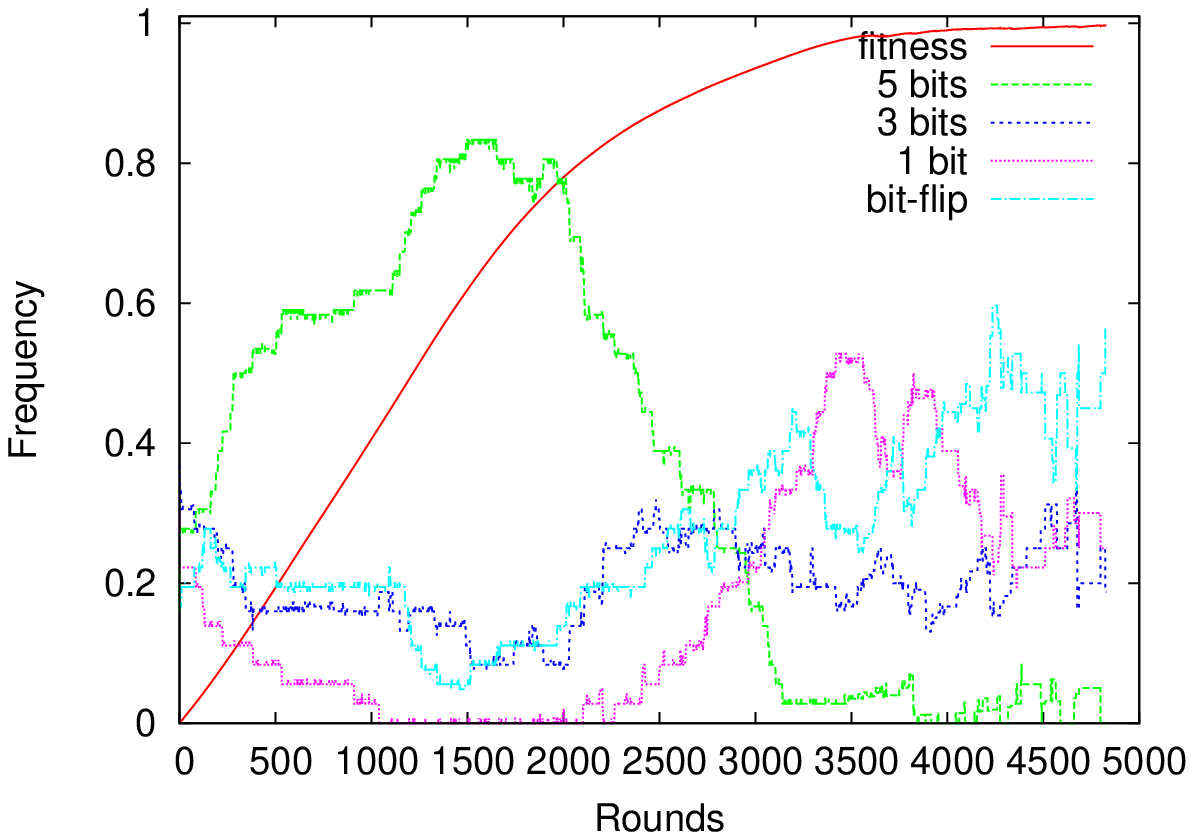} \\
\hspace{-9ex}\includegraphics[width=0.3\textwidth]{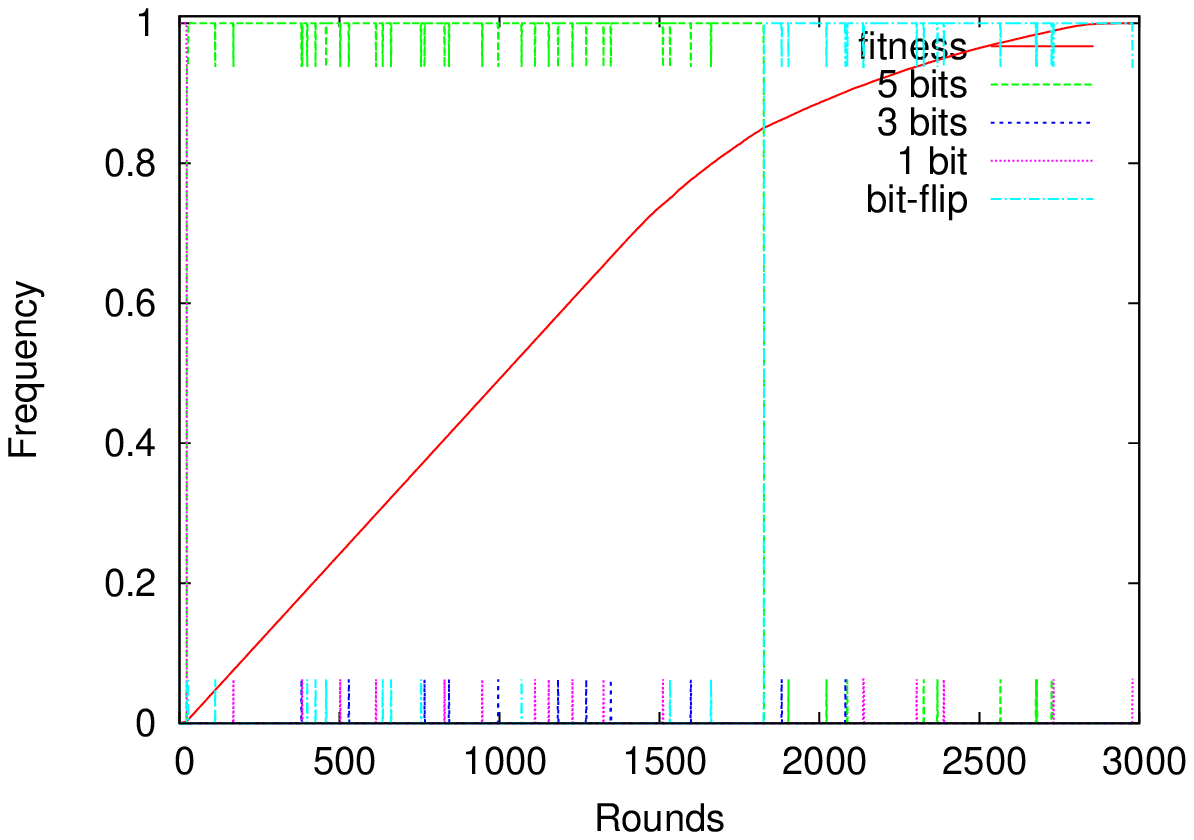} & \hspace{-5ex}\includegraphics[width=0.3\textwidth]{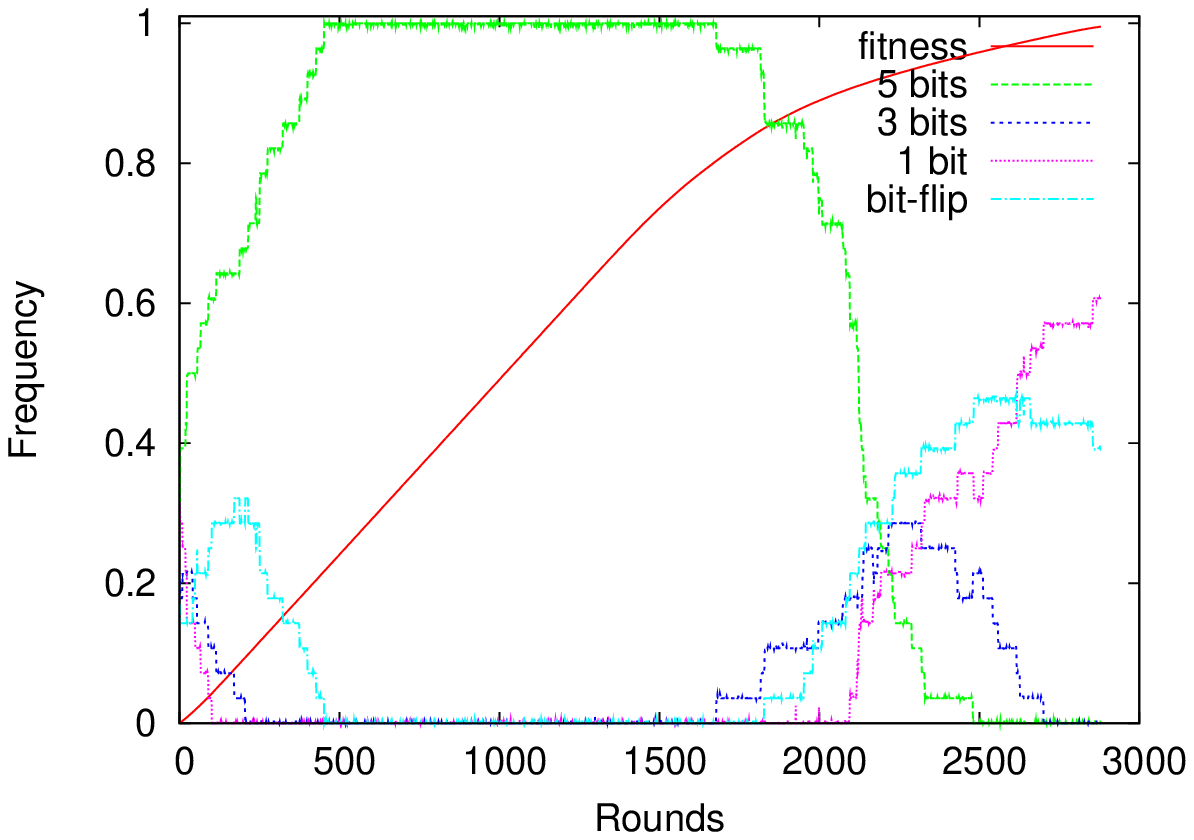} \\
\hspace{-9ex}\includegraphics[width=0.3\textwidth]{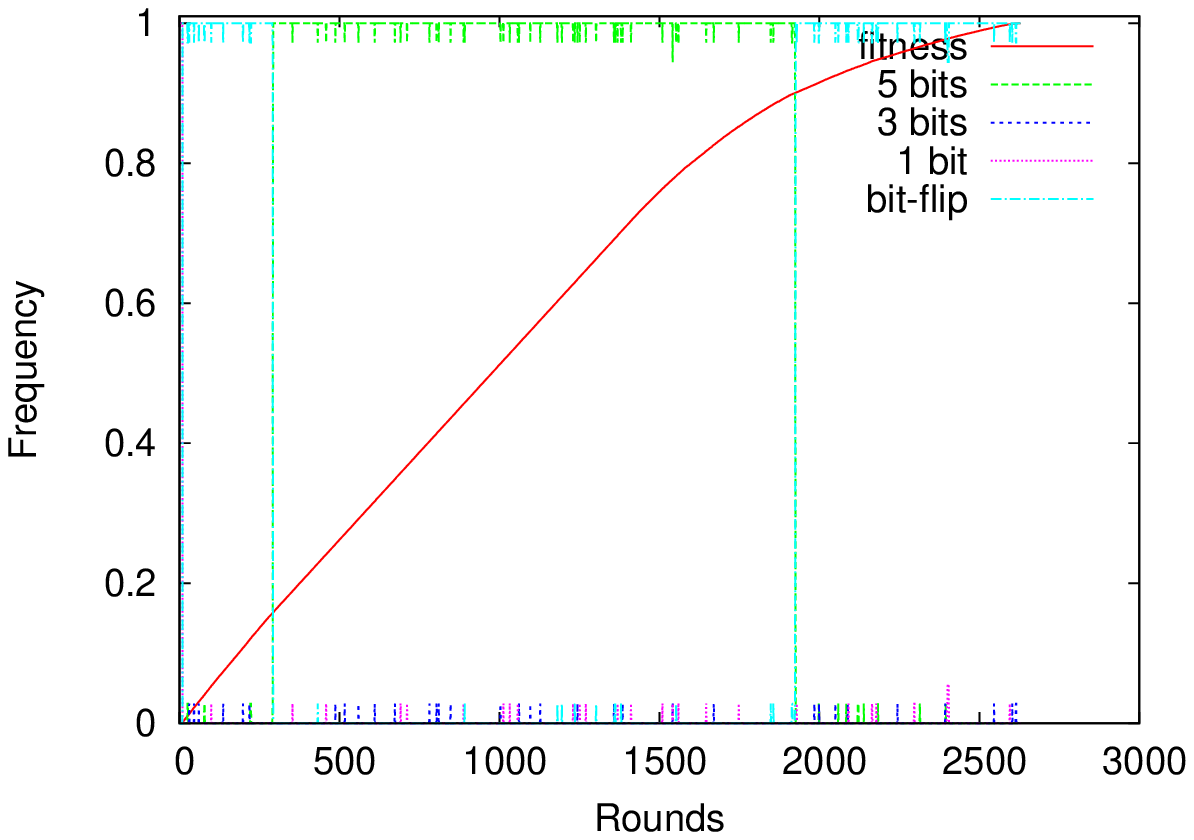} &  \hspace{-5ex}\includegraphics[width=0.3\textwidth]{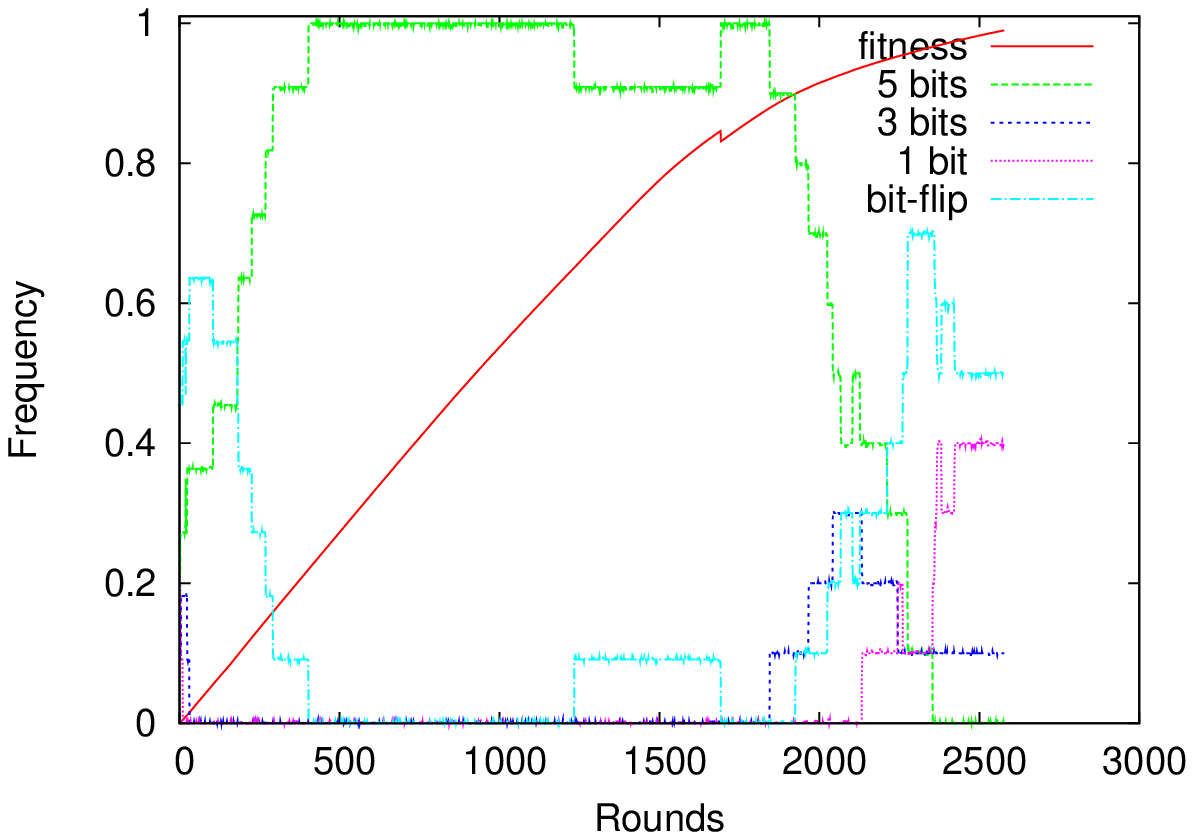} \\
\end{tabular}
\caption{Evolution of the frequency of each mutation operator for the complete network
for different sizes (from top to bottom: $n=4$, $16$, and $36$) and $p_{mut}=10^{-3}$.
Left figures are particular runs which give the median performances, 
and right figures are the average performances, as a function of the number of rounds.
\label{fig:evolnb-median-avg}}
\end{center}
\end{figure}

From Fig~\ref{fig:round-Size}, it could be surprising, at a first look, that SBM-DAMS outperforms seqOracle-DAMS, or even more the random strategy rnd-DAMS for complete topology with network size over $16$. To explain this result, Fig.~\ref{fig:evolnb-median-avg} shows the frequency of the number of nodes selecting each metaheuristic all along SBM-DAMS execution. Left figures display particular runs which obtain the median performances for different network sizes. Right figures displays corresponding average frequencies.

In left figures, we observe that nearly all nodes execute the same metaheuristic at the same time. One metaheuristic always floods the network. Only few tries of other metaheuristics appear with the rate of the metaheuristic mutation rate. Nevertheless, the population of nodes is able to switch very quickly from one operator to another one. In addition, we observe that as the number of nodes grows, the $1/l$ bit-flip mutation dominates the $5$ bits mutation at the very beginning of the runs.
This is confirmed by the average frequencies as a function of rounds (right figures) where SBM-DAMS selects bit-flip operator more frequently than $5$ bits in first rounds. At the opposite, the sequential oracle seqOracle-DAMS always chooses the $5$ bits mutation operator regardless to network size. In fact, this oracle does not care about the joint/distributed performances of other nodes and chooses the best operator from a pure sequential/selfish point of view. Hence, it always choses the $5$ bit operators since it has a better performance (in average) compared to bit-flip for low fitness. However, this selfish strategy is not optimal from the distributed point of view. In fact, let us consider $n$ nodes with initial solution $0^l$. When they all use the same $5$ bits mutation, the maximal fitness gain over the network after one round is always $5$. In that case, the fitness always increases by exactly $5$. When all nodes use the same $1/l$ bit-flip mutation, the probability that the fitness gain is strictly over $5$ is $1 - \alpha^{\lambda \cdot n}$ where $\alpha$ is the probability that at most $5$ bits are flipped\footnote{More precisely, $\alpha = \sum_{i=0}^{5} { l \choose i } (1/l)^i (1 - 1/l)^{l-i}$} in one iteration. This probability increases fast with the number of nodes. Hence, when considering the whole distributed system, the most efficient mutation operator is the bit-flip which can not be predicted by a selfish oracle. This difference between an optimal local \emph{independent} decision,  and an optimal local \emph{distributed} decision, is naturally captured by SBM-DAMS which clearly grants more bit-flip as the number of nodes increases. Note that this property is captured by SBM-DAMS without any global centralized control. In fact, while being very local and independent of the topology or any global knowledge of the network, the simple local communication policy used SBM-DAMS seems to be very efficient.

Notice also that the idea of optimal distributed decision, which is clearly demonstrated at the beginning of a run, could also appear later in the run. However, it is difficult to compare our SBM-DAMS with a global distributed oracle strategy as the possible of different of the system (for example here at most $l^n$), and the number of possible metaheuristics distribution (here $4^n$) is huge. This issue is left as an open question and addressed in the conclusion.

\subsection{Parallelism properties}
\label{sec:parallelism}

In this section, we study the performance of SBM-DAMS compared to a pure sequential adaptive strategy. Our goal is to evaluate the impact of parallelism and distributed coordination introduced by SBM-DAMS compared to a sequential setting. Let us first remark that at each round of SBM-DAMS, each node is allowed to run an iteration of a $(1+\lambda)$-EA (among four operators). Thus, over all nodes, SBM-DAMS performs $\lambda\cdot n$ evaluations in parallel. Therefore, to be fair, SBM-DAMS should be compared to a pure sequential strategy which is allowed to run a $(1+\lambda\cdot n)$-EA at each iteration. This, exactly what we report in the following experiments, where the performance of SBM-DAMS is studied for fixed $\lambda\cdot n = 50$.

\begin{figure}[ht!]
\begin{center}
\includegraphics[width=0.3\textwidth]{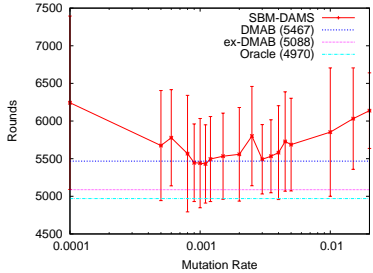}
\caption{Average of number of rounds of SBM (complete topology, $n=50$ and $\lambda=1$)
according to metaheuristic mutation rate Vs DMAB, ex-DMAB and the sequential Oracle.}
\label{fig:simple-dmab}
\end{center}
\end{figure}

In Fig~\ref{fig:simple-dmab}, we report the performance of SBM-DAMS according to the metaheuristic mutation rate $p_{mut}$ using a complete graph with $n=50$ nodes and $\lambda=1$, i.e., each node is allowed to run an iteration of a $(1+1)$-EA at each round. First, we observe as previously that SBM performs better for a mutation probability around $10^{-3}$. More importantly, for mutation rate $p_{mut}=10^{-3}$, the average number of rounds before SBM finds the best solution is $5441$ (standard deviation is $593$). As depicted in Fig~\ref{fig:simple-dmab}, SBM-DAMS is thus slightly better (in average) than the adaptive multi-armed bandit strategy (DMAB) from~\cite{Fialho08}, but worse than the oracle strategy and the best existing multi-armed bandit strategy from \cite{Fialho10}. However, we remark that DMAB algorithms are different in nature from SBM. In fact, DMAB is allowed to update the information about mutation operators each time a new individual (among $\lambda$) is generated during the $(1+\lambda)$-EA iteration, whereas SBM is not since the $(1+1)$-EA in SBM-DAMS is executed in parallel by all nodes. Overall, we can yet reasonably state that SBM is competitive compared to other sequential strategies in terms of number of rounds needed to reach to optimal solution. More importantly, since SBM is distributed/parallel in nature, an effective implementation of SBM on a physical network can lead to better running time compared to DMAB which is inherently sequential\footnote{Each time an individual (over $\lambda$) is generated by DMAB, the reward information about mutation operators needs to be updated.}. In other words, the effective running time of SBM-DAMS could be divided by $\lambda=50$ (compared to DMAB) which is the best one can hope to obtain when using $(1+\lambda)$-EAs.

\begin{figure}[ht!]
\begin{center}
\begin{tabular}{cc}
\hspace{-8ex}\includegraphics[width=0.3\textwidth]{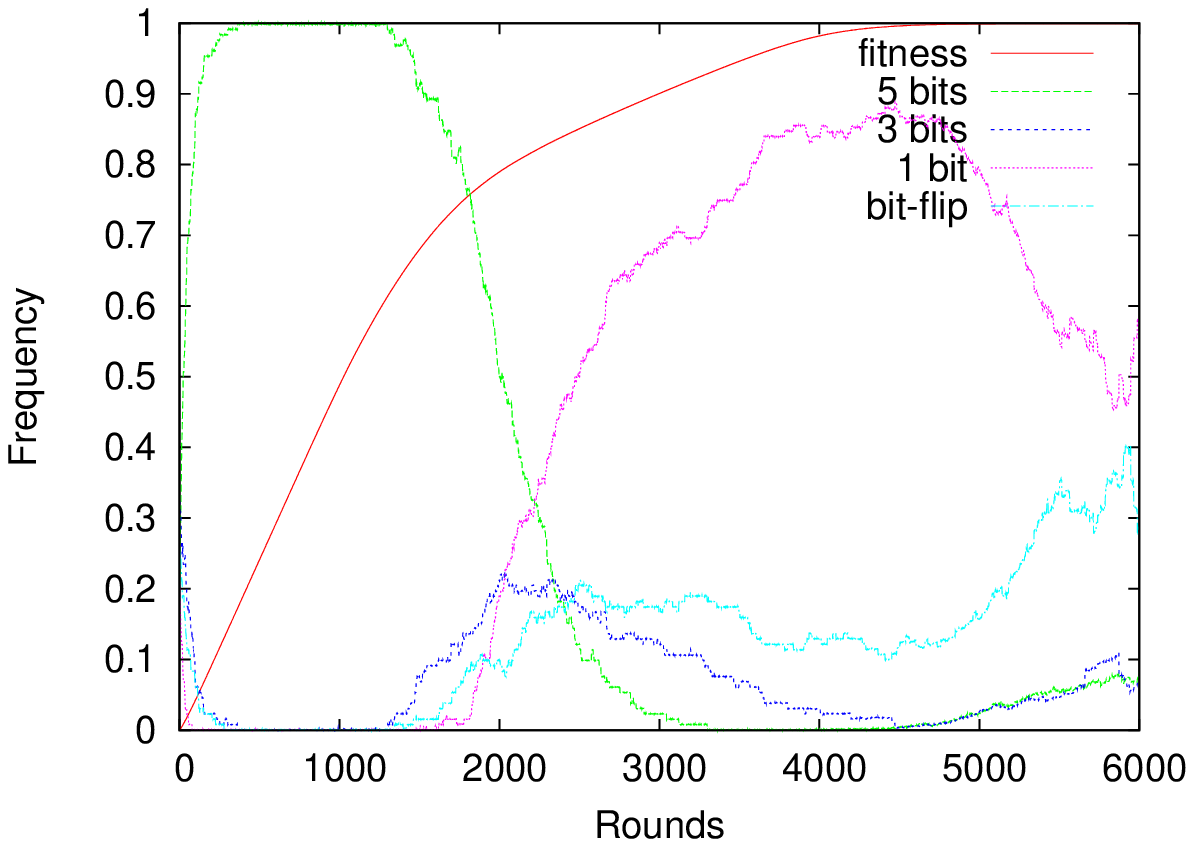} & \hspace{-5ex} \includegraphics[width=0.3\textwidth]{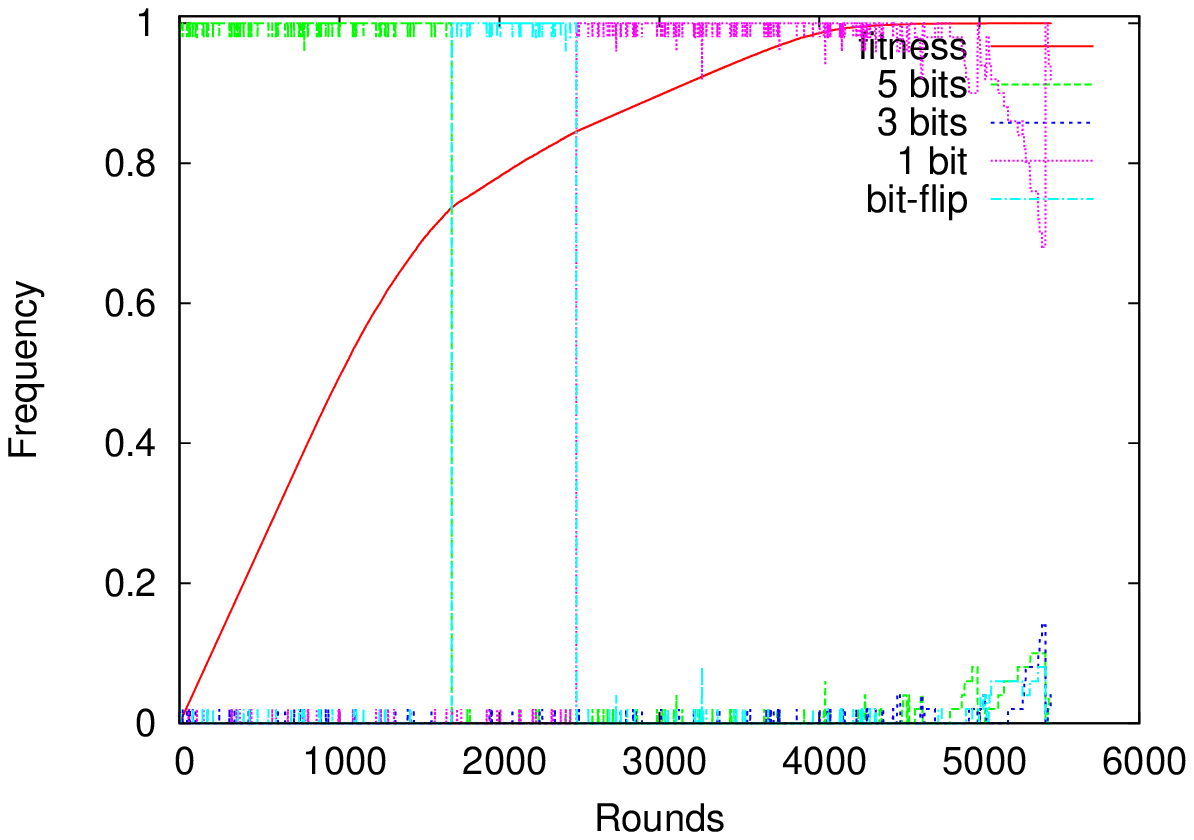}\\ 
\multicolumn{2}{c}{\hspace{-8ex}\includegraphics[width=0.3\textwidth]{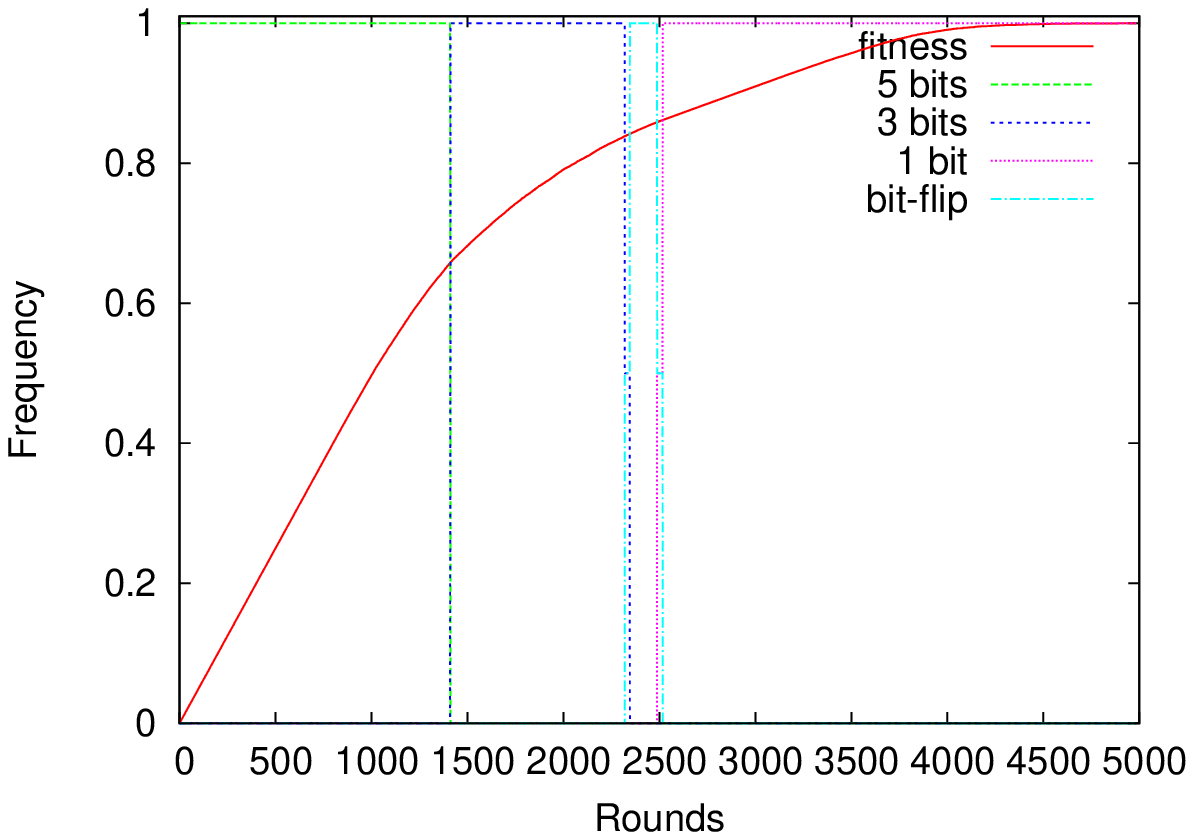}}\\
\end{tabular}
\caption{Top left (resp. right): Evolution of the average number of nodes (resp. number of nodes for the median run) of each mutation operator using a complete topology with $n=50$, $\lambda=1$, and $p_{mut}=10^{-3}$. Bottom: Evolution of the mutation operator in average according the the sequential oracle with $\lambda=50$.
\label{fig:simple-evol}}
\end{center}
\end{figure}

In Fig~\ref{fig:simple-evol}, we report the adaptation properties of SBM-DAMS compared to the sequential oracle strategy. Top left figure shows that nodes act globally in average closely to the sequential oracle illustrated by bottom figure. In particular, one can see that nodes switch (in average) to the $3$-bit and the rate-bit mutation operators around the same round than the sequential oracle. This is an interesting property especially for these two operators which should be used during a small window of rounds to speed up the search. Top right figure displays one particular run which obtains the median performance over all runs. Again, one can clearly see that all nodes can switch from a mutation operator to a better one very quickly at runtime.

\begin{figure}[ht!]
\begin{center}
\begin{tabular}{cc}
\hspace{-8ex}\includegraphics[width=0.3\textwidth]{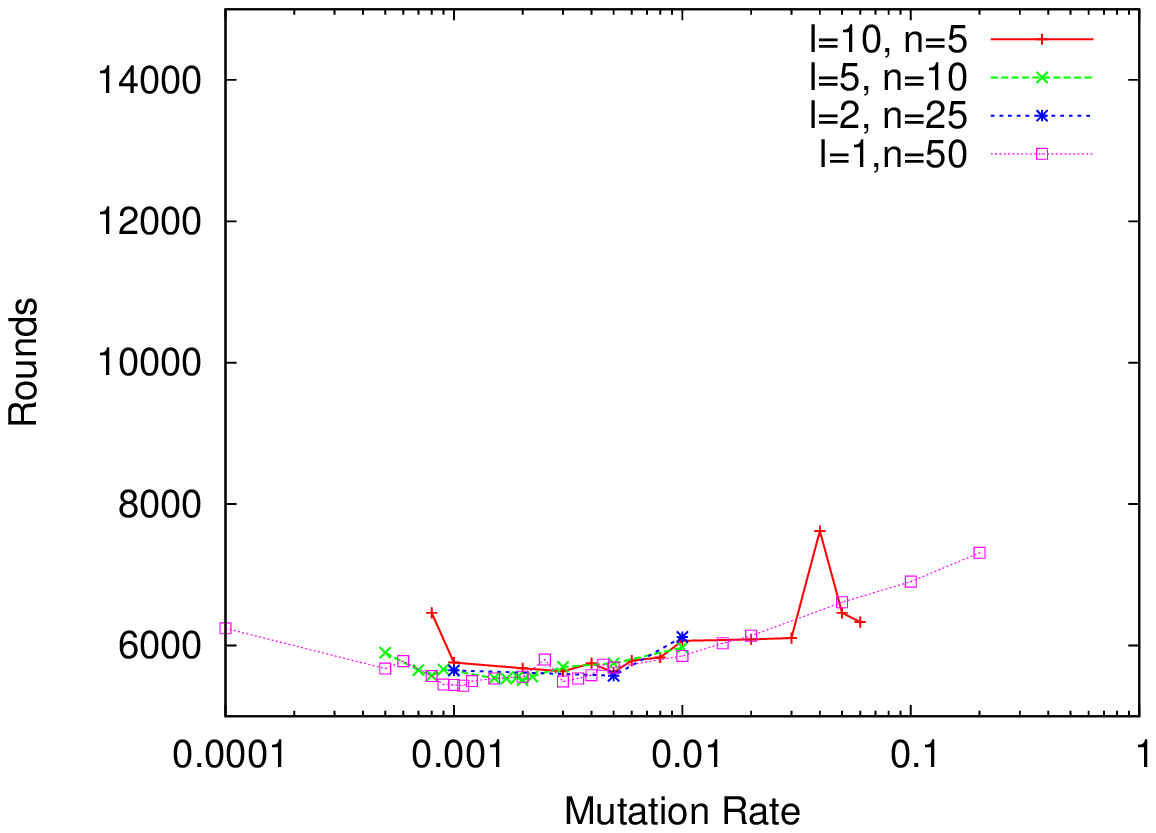} & \hspace{-5ex}\includegraphics[width=0.3\textwidth]{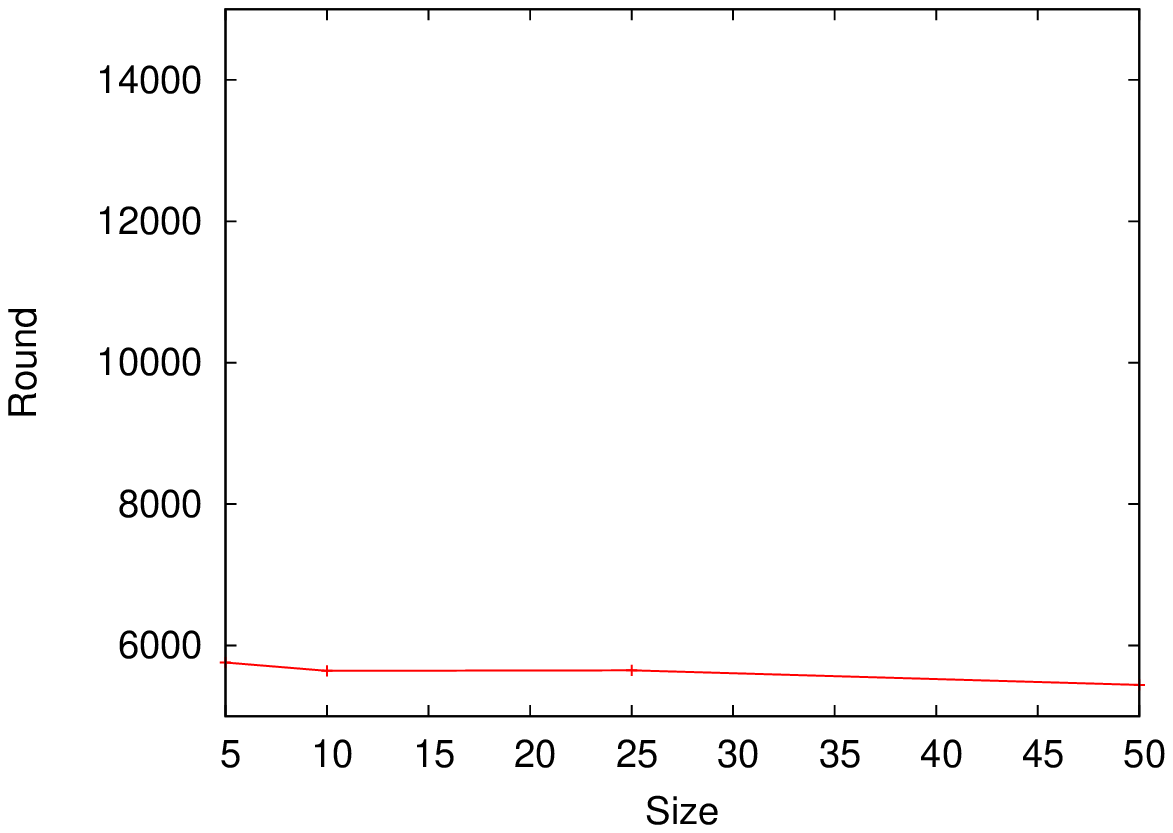}\\

\hspace{-8ex}\includegraphics[width=0.3\textwidth]{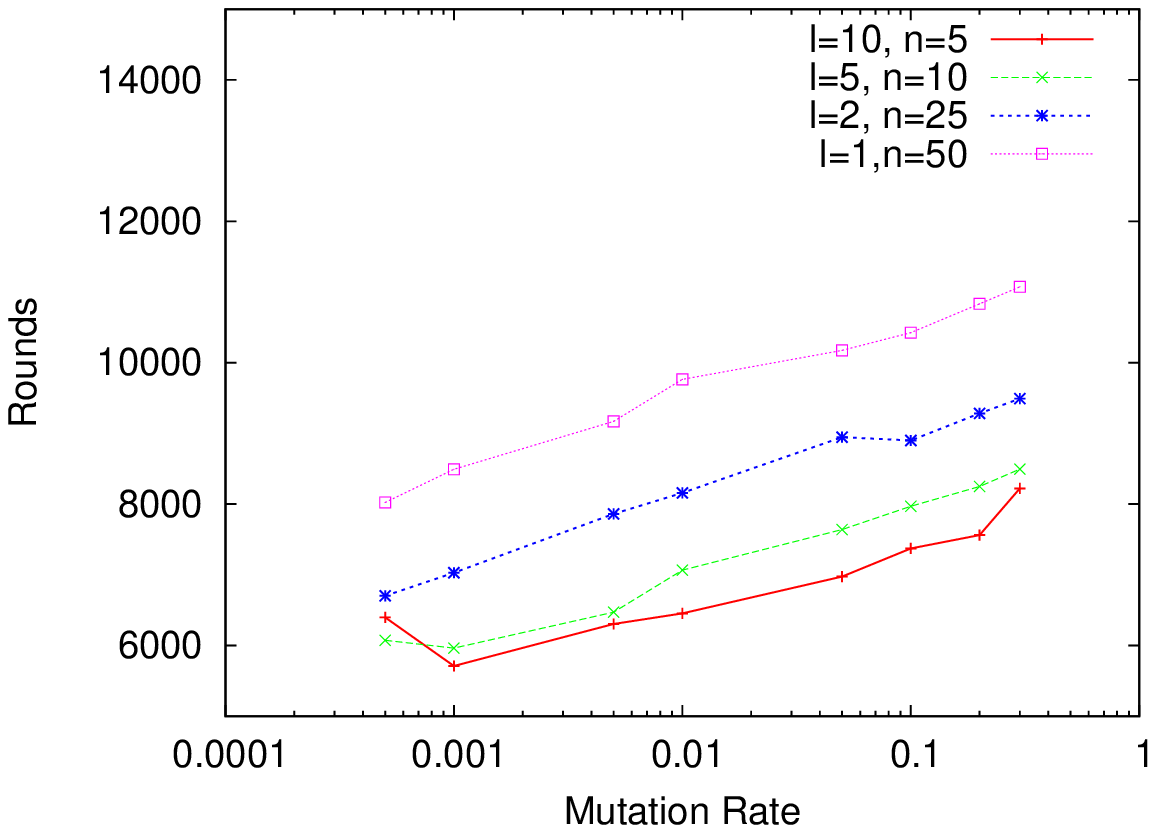} & \hspace{-5ex}\includegraphics[width=0.3\textwidth]{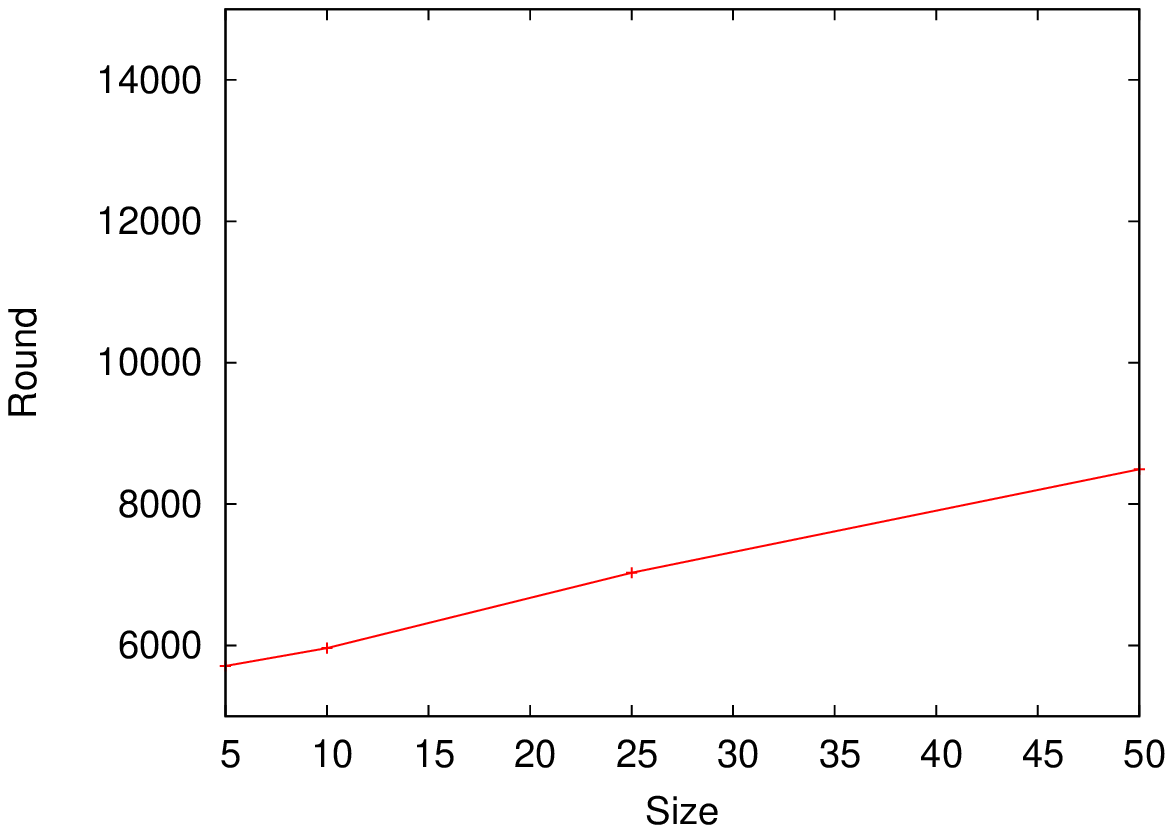}\\

\hspace{-8ex} \includegraphics[width=0.3\textwidth]{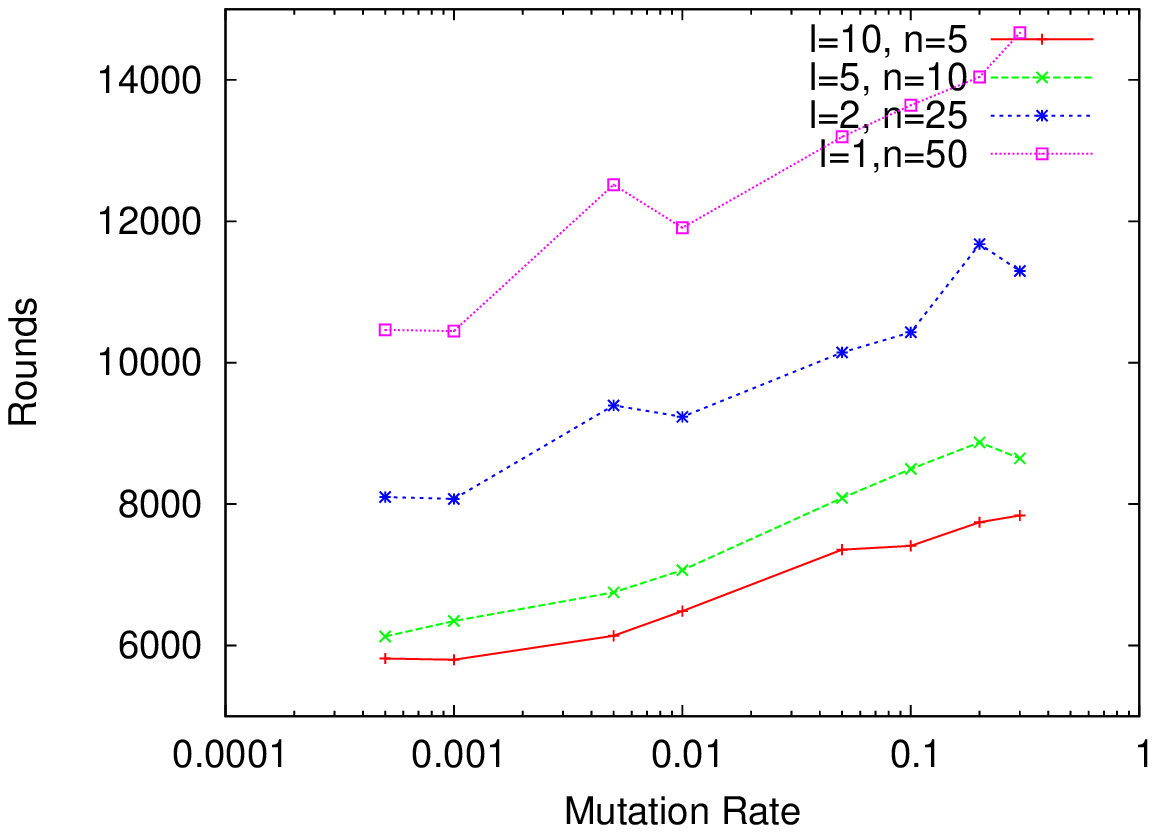} & \hspace{-5ex} \includegraphics[width=0.3\textwidth]{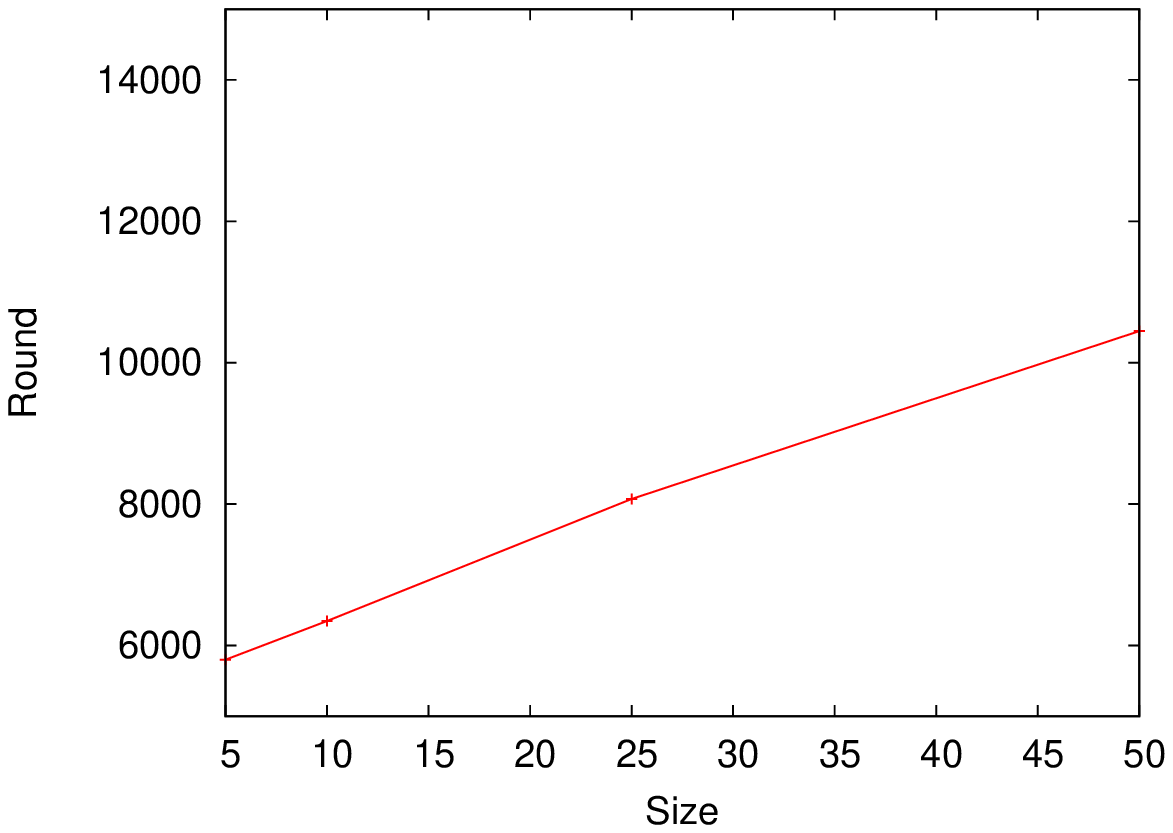}
\end{tabular}
\caption{ From top to down: complete, grid, cycle topologies. Left: Average number of rounds
according to metaheuristic mutation rate. Right: Average number of rounds
according to size $n$ verifying $\lambda\cdot n = 50$.
\label{fig:round-mutation}}
\end{center}
\end{figure}

Finally, in Fig~\ref{fig:round-mutation}, we report the performance of SBM-DAMS for different topologies using different sizes $n$ such that $\lambda\cdot n =50$, namely $(\lambda,n) \in \set{(10,5),(5,10),(2,25),(1,50)}$. Left figures reports the robustness of SBM-DAMS according to mutation probability $p_{mut}$. Right figures displays the evolution of the number of rounds for a mutation parameter $p_{mut}=10^{-3}$ as a function of $n$. One can see that for the complete topology, SBM-DAMS is robust and its performances is almost constant for all $l\cdot n =50$. However, it is without surprise that for the grid and the cycle, SBM-DAMS performances degrade as the size of the network goes higher (equivalently as $\lambda$ goes smaller). This can be intuitively explained as following: (i) for smaller $\lambda$ the probability that a given mutation operator produces a better solution is smaller and (ii) the probability that a node, which has mutated from one operator to a better one, informs all other nodes gets smaller as the network diameter gets higher. Hence, from a pure parallel point of view, the considered topology could have a non negligible impact on the overall speedup of SBM-DAMS as previously suggested by Fig.~\ref{fig:round-Size}.

\section{Conclusion and open questions}
\label{sec:conc}

In this paper, we have proposed a new distributed adaptive method, the select best and mutate strategy (SBM), in the generic framework of distributed adaptive metaheuristic selection (DAMS) based on a three-layer architecture.
SBM strategy selects locally the metaheuristic with the best instant reward and mutates it according to a mutation rate parameter. 
The conducted experimental study shows SBM robustness and efficiency compared to naive distributed and sequential strategies. Many futures studies and open question are suggested by DAMS. Firstly, it would be interesting to study DAMS for other problems and using other EAs. For instance, when considering population-based metaheuristics many issues could be addressed for population migration and population update. In particular, one can use some distributed adaptive policies to guide population evolution. Adaptive migration taking into account the performance of crossover operators should also be introduced and studied. DAMS could also be particularly accurate to derive new algorithms to tackle multi-objective optimization problems, since, for instance, one can use the parallelism properties of DAMS to explore different regions of the search space by adapting the search process according to different objectives.

For SBM-DAMS, many questions remain open. For instance, one may ask at what extent one can design a distributed strategy based on multi-armed bandits techniques to select metaheuristics, compute metaheuristic rewards, and adapt the search accordingly. Further work needs also to be conducted to study the impact of the distributed environment. For instance, asynchrony and low level distributed issues on concrete high performance and large scale computing platforms, such as grid and parallel machines, should be conducted. Another challenging issue is to adapt local communications between nodes, for instance, to use the minimum number of messages and obtain the maximum performances. In particular, it is not clear how to tune the number of nodes and the topology in a dynamic way in order to balance the communication cost and the computation time, i.e., parallel efficiency. From the theoretical side, we also plan to study the parallelism of SBM analytically and its relation to a fully distributed oracle that still needs to be defined.

\vfill\eject

\bibliographystyle{abbrv}


\balancecolumns

\end{document}